\documentclass[letterpaper]{article} % DO NOT CHANGE THIS
\usepackage{aaai2026}  % DO NOT CHANGE THIS
\usepackage{times}  % DO NOT CHANGE THIS
\usepackage{helvet}  % DO NOT CHANGE THIS
\usepackage{courier}  % DO NOT CHANGE THIS
\usepackage[hyphens]{url}  % DO NOT CHANGE THIS
\usepackage{graphicx} % DO NOT CHANGE THIS
\usepackage{natbib}  % DO NOT CHANGE THIS AND DO NOT ADD ANY OPTIONS TO IT
\usepackage{caption} % DO NOT CHANGE THIS AND DO NOT ADD ANY OPTIONS TO IT
\usepackage{algorithm}
\usepackage{algorithmic}
\usepackage{tabularx}
\usepackage{graphicx}
\usepackage{amsmath}
\usepackage{amsfonts}       % blackboard math symbols
\usepackage{newfloat}
\usepackage{listings}
\DeclareCaptionStyle{ruled}{labelfont=normalfont,labelsep=colon,strut=off} % DO NOT CHANGE THIS
\floatstyle{ruled}
\newfloat{listing}{tb}{lst}{}
\floatname{listing}{Listing}
\title{Neurosymbolic LoRA: Why and When to Tune Weights vs. Rewrite Prompts}

\author{
  Kevin Wang\equalcontrib, Neel P. Bhatt\equalcontrib, Cong Liu\equalcontrib, Junbo Li, Runjin Chen, Yihan Xi, Timothy Barclay, Alvaro Velasquez, Ufuk Topcu, Zhangyang Wang}
   
\usepackage{bibentry}
\begin{document}

\maketitle

\begin{abstract}
Large language models (LLMs) can be adapted either through \emph{numerical} updates that alter model parameters or \emph{symbolic} manipulations that work on discrete prompts or logical constraints. While numerical fine-tuning excels at injecting new factual knowledge, symbolic updates offer flexible control of style and alignment without retraining. We introduce a neurosymbolic LoRA framework that dynamically combines these two complementary strategies. Specifically, we present a unified monitoring signal and a reward-based classifier to decide when to employ LoRA for deeper factual reconstruction and when to apply TextGrad for token-level edits. Our approach remains memory-efficient by offloading the symbolic transformations to an external LLM only when needed. Additionally, the refined prompts produced during symbolic editing serve as high-quality, reusable training data, an important benefit in data-scarce domains like mathematical reasoning. Extensive experiments across multiple LLM backbones show that neurosymbolic LoRA consistently outperforms purely numerical or purely symbolic baselines, demonstrating superior adaptability and improved performance. Our findings highlight the value of interleaving numerical and symbolic updates to unlock a new level of versatility in language model fine-tuning.

% \textcolor{orange}d
% {Edit version: Low-Rank Adaptation (LoRA) has demonstrated impressive performance in fine-tuning large language models (LLMs) with low memory overhead. However, purely numerical parameter updates often struggle with closing the gap on due to the local restriction of gradient-based method. To address this, we propose \textbf{Neurosymbolic LoRA}, a hybrid framework that integrates LoRA's efficient learning with symbolic updates to enforce constraints and refine behavior. By leveraging LoRA neuron signals or a reward model-based classifier, the framework dynamically switches between update modes. Experiments across diverse tasks suggests consistent improvements. We additionally analyze the interplay between numeric and symbolic components, demonstrating Neurosymbolic LoRA can further facilitate the creation of high-quality, transferable datasets, thereby enhancing fine-tuning for new models.}:

\end{abstract}

\section{Introduction}
\label{sec:intro}

Neurosymbolic artificial intelligence (AI) \cite{garcez2023neurosymbolic} offers a way forward by combining the raw pattern-recognition prowess of large language models (LLMs) with the structured constraints of symbolic reasoning. Taking LLM fine-tuning or performance adaptation---presumably the most popular application case in practice---for example: numerical approaches such as Low-Rank Adaptation \cite{hu2021lora} (LoRA) dominate the practical pursuit of efficient adaptation: they modify model parameters numerically, injecting factual knowledge or bridging domain gaps. However, purely parametric updates can struggle with tasks demanding precise logical constraints, interpretability, or strict style alignment \cite{yang2022injecting, li2024neuro}. 

Conversely, symbolic techniques---in a broad sense---can manipulate discrete tokens, prompts, or logic-based decoding rules to steer a model’s behavior and enforce constraints without altering the underlying parameters. Earlier representative works include prompt tuning and prefix tuning \cite{shin2020autoprompt,li2021prefix,lester2021power,liu2021p,zhang2021differentiable,khattab2022demonstrate, ccoplu2024prompt}. Latest variants of prompt tuning such as DSPy \cite{khattab2024dspy} and TextGrad \cite{yuksekgonul2024textgrad} apply gradient-like signals to \emph{edit} prompts, thereby indirectly shaping the model’s internal activations by making token-level edits. These approaches stand as instances of a broader class of “symbolic” approaches---one could instead apply logic rules \cite{lu2020neurologic}, finite-state automata \cite{yang2024fine}, domain verifiers \cite{kambhampati2024llms}, or other structured constraints during generation. 

\begin{figure*}[t]
    \centering
    \includegraphics[width=\linewidth]{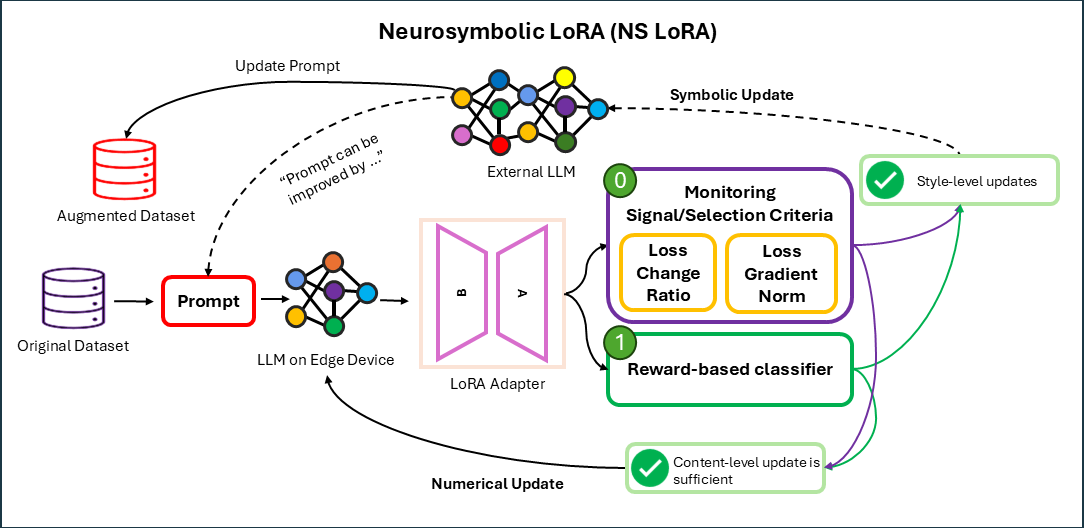}
    \caption{Overview of the neurosymbolic LoRA framework: we visually depict the workflow for our framework. In particular, there are two primary means to alternate between numerical and symbolic updates, either via a selection criteria that utilizes a training signal or using a reward-based classifier.}
    \label{fig: framework}
    % \vspace{-5pt}
\end{figure*}

It is precisely in the interplay between numerical parameter updates and symbolic manipulation of discrete structures that we see \textit{complementary strengths}. Numerical adaptations (e.g., via LoRA) excel at fully integrating new domain knowledge, especially for “content-centric” fixes or when bridging factual gaps. Symbolic manipulation shines at style alignment, interpretability, and discrete constraint enforcement. When used in isolation, each has its limitations; combined judiciously, they can yield a more robust, interpretable, and controllable system. Recent work  \cite{hoang2023dynamic} offers illuminating evidence about the differences between re-training (numerical fine-tuning) and prompting-based approaches (symbolic) in recovering the performance of compressed LLMs.  Their findings reveal an intriguing divergence in how these methods reshape the internal dynamics of a model. Compared to LoRA, the attention mechanism arising from prompting exhibits marked deviations from the baseline (i.e., the pre-compressed model), suggesting that prompting triggers a form of \emph{contextual redirection}. This stands in contrast to the activation patterns, which align more closely with LoRA. Notably, LoRA incorporates residual connections at every layer to preserve congruity with the original model, whereas prompting modifies only the input component. Given the distinct mechanisms between these two methods, it is unexpected to observe a high degree of similarity in their activation patterns.

A notable recent work \cite{soylu2024sftprompt} proposed to alternate between fine-tuning and prompt optimization independently over the same dataset after a fixed number of epochs. It showed noticeable accuracy gains compared to applying only one optimization technique. Although the simple condition triggering the switch could result in unnecessarily expensive prompt optimization, it showed a promising direction of combining the two optimization techniques.

These observations collectively imply that prompting-based approaches (``symbolic")---by focusing on attention-level redirection---can effectively tap into \emph{latent knowledge} still embedded in the compressed model’s internal representations. In other words, symbolic updates target style alignment rather than factual inaccuracies and, hence, demonstrate complementary benefits to numerical reconstruction (LoRA), which focuses on re-injecting factual knowledge from external data. The observed performance gains in \cite{hoang2023dynamic} indicate that prompting-based redirection is surprisingly efficient for knowledge-related tasks, pointing to a synergy between parametric fine-tuning and prompt manipulation: their joint optimization can dynamically exploit internal representations for stylistic or factual changes. 

However, traditional prompt-tuning often relies on gradient signals within the same LLM, rendering prompt edits relatively local and constrained in both location and length \cite{wang2024universality}. Moreover, these updates remain tightly coupled to the model’s existing parameters. In contrast, newer prompt-editing methods integrate \textbf{external oracles} in the form of \textit{auxiliary LLMs or other symbolic engines} to perform discrete token-level changes. This effectively brings in an independent source of knowledge, beyond what the target model’s gradients alone can convey, and grants more free-form editing capacity (rather than fixed-position or fixed-length prompts). Techniques like TextGrad exemplify this broader flexibility, enabling more global, wide-ranging changes to the prompt. These advances reinforce the broader theme that numerical and symbolic updates are not mutually exclusive but rather complementary levers for constructing more robust and adaptive LLMs.

\subsection{Our Contributions} 

Building on the broader aim of \emph{adaptively} interleaving numerical and symbolic strategies, we instantiate our approach using LoRA \cite{hu2021lora} as the numerical component and TextGrad \cite{yuksekgonul2024textgrad} as the symbolic component. LoRA is a natural choice for memory-efficient fine-tuning, excelling at parameter-level updates that inject or correct factual knowledge. TextGrad, on the other hand, serves as an exemplar of token-level editing using an external LLM as a gradient oracle to re-write prompts, thus representing a flexible, discrete steering mechanism. Despite focusing on LoRA and TextGrad in this paper, our overarching methodology is not constrained to these two techniques. It serves as a more general \emph{neurosymbolic} template where different numerical approaches (e.g., full fine-tuning, low-rank adapters, or quantization) and diverse symbolic algorithms (e.g., logic-based constraints, chain-of-thought expansions, or finite automata) can be seamlessly integrated.

Critically, this is \emph{not} a naïve “brute force” layering of LoRA and TextGrad. Rather, we propose a structured feedback loop wherein each approach is invoked under specific conditions. Numerical LoRA updates perform deeper factual reconstruction if the model consistently fails content checks, while symbolic edits (TextGrad) handle stylistic or local re-routing tasks that do not require wholesale re-injection of knowledge. These updates are triggered either by a monitoring signal that tracks performance changes across epochs or by a reward-based classifier that detects whether the input prompt is style-driven or content-oriented. We thus avoid an uncoordinated mix-and-match of techniques, ensuring that each algorithm is used precisely where it is most effective. Given LoRA’s fundamental role in parameter-level adaptation, we name our unified approach “\textbf{neurosymbolic LoRA}” (see Figure \ref{fig: framework}), reflecting the synergy between the numerical (low-rank parameter) and symbolic (prompt editing) dimensions.

Despite involving external LLM calls, our neurosymbolic pipeline remains \textbf{highly practical} for real-world contexts:
\begin{itemize}
    \item First, outsourcing the symbolic updates (e.g., TextGrad prompt rewriting) to an external API can substantially \emph{reduce on-device memory consumption}, trading off local memory usage for latency. This is appealing in resource-constrained or edge deployments, where the cost of storing additional modules is prohibitive. 
    \item Second, we deliberately structure our approach to invoke external symbolic edits \emph{sparsely}, only when a monitoring or classification signal indicates that parametric fine-tuning has plateaued or that style realignment is essential. This ensures that latency overhead does not accumulate excessively. 
    \item Finally, a notable benefit emerges from our symbolic editing procedure: it “re-makes” or transforms the training data (or generated training examples) into forms more consumable by the target LLM. In data-scarce domains—such as mathematical reasoning, biomedical text, or legal arguments—collecting high-quality, annotated data can be extremely costly. Our symbolic rewriting, driven by an external LLM, can yield \emph{reusable, higher-quality training sets} for downstream or future models, providing a lasting resource beyond the immediate fine-tuning task.
\end{itemize}

Extensive experiments demonstrate the effectiveness of our neurosymbolic LoRA approach in enhancing LLM capabilities, particularly with limited annotated data or in tasks that demand rapid adaptation to evolving requirements. This interwoven strategy significantly boosts both factual fidelity and stylistic control, while remaining mindful of practical constraints like memory, latency, and data annotation costs. 

In summary, our technical contributions are outlined below:
\begin{itemize}
    \item \textbf{Unified Monitoring Signal for Numerical and Symbolic Updates.} We propose a mechanism that tracks performance indicators, such as loss changes across epochs, to determine when to apply numerical versus symbolic updates. If numerical fine-tuning (e.g., LoRA) saturates, symbolic rewriting (e.g., TextGrad) is triggered to refine style or re-route existing knowledge; likewise, if symbolic edits yield limited gains, the framework reverts to numerical adaptation. This principled switching leverages the strengths of both methods without resorting to a brute-force combination, and aligns with prior evidence that knowledge may remain “displaced” rather than deleted in compressed models \cite{hoang2023dynamic}.
    \item \textbf{Reward-Based Classifier.} To complement the monitoring signal, we develop a reward-based classifier that identifies whether an input primarily requires factual (content-level) updates or style/constraint alignment. This yields a more context-sensitive decision boundary for switching between parametric and symbolic manipulations.
\item \textbf{Practical Integration with External LLMs and Data Remaking.} Our approach remains feasible in memory-constrained environments by calling an external LLM for symbolic edits only when needed, thus offloading heavy computations. Moreover, the prompt transformations generated by our framework can serve as \emph{re-made} training data, which is especially beneficial in specialized or data-scarce domains (e.g., math QA). These augmented datasets can be reused to train subsequent models, reinforcing both cost-efficiency and performance gains over time.
    \item \textbf{Empirical Validation and Scalability.} We extensively evaluate the proposed hybrid framework across multiple LLM backbones and tasks. Our results show that the synergy of numerical (LoRA) and symbolic (TextGrad) updates outperforms purely numerical or purely symbolic methods, especially in low-data settings. The framework also offers a scalable blueprint for integrating a variety of numerical-symbolic pairs beyond LoRA and TextGrad, opening avenues for broader neurosymbolic AI research.
    % potential \item Neurosymbolic is orthogonal to may prior fine-tune method and can be combined with many of the them such as dpo, kto...?
\end{itemize}

\section{Related Works}
\label{sec:realted_works}

There has been a growing trend towards the synthesis of numerical and symbolic approaches to align the outputs of LLMs to certain task requirements. Fine-tuning enables the alteration of internal model parameters that have been pre-trained to improve their performance on a particular task. There is a wealth of research in this area showcasing the relevance of fine-tuning pre-trained models in specific domains such as demonstrated by \cite{yang2023fine, bhatt2024know}. These works demonstrate the enhancement of factual accuracy through injecting new knowledge into the language models. In addition, due to the increasing need to deploy models on edge devices, recent works such as by \cite{lora, xu2023parameter, li2021prefix, lester2021power} present memory-efficient fine-tuning methods aim at reducing the computational burden in fine-tuning. Specifically, LoRA~\cite{lora} applies low-rank decomposition to update only a small subset of weights offering efficient storage of swapable adapters for specific tasks.

Prompt optimization is another important aspect of LLM alignment, which deals with the improvement of input prompts to obtain more accurate responses without changing the internal model parameters. Techniques such as DSPy \cite{khattab2023dspy, khattab2024dspy} have emerged as important tools that provide automated methods for prompt optimization. Similarly, TextGrad \cite{yuksekgonul2024textgrad} introduces gradient-based methods to iteratively enhance prompts, thus ensuring better alignment with desired outputs. These methods demonstrate that prompt engineering serves as an alternative to standard model fine-tuning, emphasizing stylistic control over language model outputs rather than the injection of factual knowledge. Soft prompt tuning~\cite{lester2021power} is another method that learns continuous embeddings for task-specific prompts. Prefix-tuning~\cite{li2021prefix} extends soft prompt tuning by pre-pending continuous task-specific vectors to the input sequence, enabling the model to condition on task-specific representations without modifying its core parameters. Unlike discrete prompt tuning, where prompts are manually designed, prefix-tuning optimizes a learned prefix that serves as virtual tokens attended by a transformer. 

Finally, symbolic methods for LLM adaptation add another layer of sophistication by encoding structured reasoning and logical constraints into model behavior. Recent works have explored the distillation of symbolic representations from the outputs of language models and subsequently utilize feedback from formal verification for automated fine-tuning \cite{yang2024fine}. For instance, a recent work \cite{west2022symbolic} distills common sense knowledge from GPT-3 into a knowledge graph and multiple neural models that are two orders of magnitude lighter and yet achieve superior commonsense reasoning performance. Similarly, another framework coined Ctrl-G \cite{zhang2024adaptable} takes symbolic logic constraints as inputs and integrates them into the distillation of a small language model (SLM) such that the generated text adheres to these constraints. These methods uplift neural representations into a symbolic space and obtain a measure of compliance with task descriptions.

By harnessing the complementary advantages of numeric and symbolic methodologies, our framework ensures not only factual alignment of generated text but also enforces style alignment. Moreover, it enables the reuse of high-quality prompts for future training, especially in data-scarce domains such as mathematical reasoning.

\section{Methodology}
% \subsection{Preliminaries}
% \label{sec:preliminaries}

%\subsubsection{Problem Statement}
Fine-tuning LLMs, especially using LoRA, is generally performed in memory-constrained environments and on edge devices. Performing symbolic prompt optimization via TextGrad, for instance, is comparatively resource intensive given the need to host both a teacher and student model. Consequently, co-hosting LoRA and TextGrad on the same edge device in such memory-constrained settings is infeasible without a tradeoff on memory with latency. Accordingly, TextGrad can be hosted via a cloud computing resource and accessed via an external API call.

\subsection{Seeking Training Signals for Switching Between Numerical and Symbolic Updates}
\label{sec: monitoring_signals}

We introduce a framework that adaptively governs the transition between numerical and symbolic updates in model fine-tuning. At the heart of our framework is a unified monitoring signal correlated with performance changes in the model. This signal, derived from key performance indicators such as training loss change ratio or gradient norm, provides a continuous assessment of model performance at each epoch. By tracking when this monitoring signal saturates while performing numerical updates, a trigger to a symbolic update phase can be initiated to refine style or re-route internal representations, echoing the concept of “knowledge displacement” by allowing latent knowledge to be reclaimed or redirected efficiently. 

An ideal selection of the monitoring signal should ultimately inform whether it is worthwhile for new factual knowledge to be injected (via numerical updates) using the input sample or if contextual redirection for style alignment (via symbolic updates) is required.

At the start of model training, we nominally initiate the \textbf{numerical update} phase using LoRA in an attempt to learn new factual knowledge from the training dataset $\mathcal{D}$. Consider a question and answer problem setting which involves a textual description of the system prompt $P$, a question $Q$ that the model $f_\theta$ attempts to answer, a ground truth answer $A^*$, and the model's prediction $A$. We want to identify a signal $C(x)$ such that the saturation in the signal serves as a trigger to a \textbf{sparse symbolic update} phase. Aligned with our problem setting, we perform symbolic updates until the model predicts the correct answer or if the number of allowable symbolic update iterations have been reached. Subsequently, we alternate back to the numerical update phase as this is our nominal and preferred update mode. Additionally, we incorporate an optional \textbf{system prompt learning} procedure. In this stage, selected training samples contribute to a symbolic gradient signal used to update the system prompt globally at the end of each epoch. The updated system prompt is then employed in the subsequent epoch, enabling the model to benefit from accumulated symbolic insights throughout training. \vspace{-1em}

\noindent \begin{minipage}{0.45\textwidth}
\begin{algorithm}[H]
    \caption{Neurosymbolic LoRA}
    \begin{algorithmic}[1]
    \label{alg: neurosymbolic_lora}
        \STATE {\bfseries Input: } Neural network $f_\theta$, dataset $\mathcal{D}$ containing text sample $x \in \{P,Q,A\}$, ground truth answer $A^*$, criteria function $C(x)$, symbolic update ratio $p\in[0,1]$.
        \FOR{epoch}
            \STATE {\textcolor{blue}{\# Numerical update}}
            \FOR{every batch in $\mathcal{D}$}
                \STATE Update model $f_\theta$ using LoRA.
                \STATE Record criteria $C(x)$ for each sample $x$ in batch.
            \ENDFOR
            \IF{criteria $C(x)$ is not met, i.e. performance saturation has occurred under numerical update}
            \STATE {\textcolor{blue}{\# Symbolic update}}
            \STATE Select $p$ fraction in $\mathcal{D}$ with the lowest $C(x)$ and re-write $Q$ for each using \textit{TextGrad} and/or $P$.
            \ENDIF
        \ENDFOR
    \end{algorithmic}
\end{algorithm}
\end{minipage}
% \end{wrapfigure}
\vspace{0.5em}

The first unified signal we utilize to track saturation due to numerical updates is the loss change ratio across consecutive epochs, defined as \( C(x)_1 = \frac{\mathcal{L}_t - \mathcal{L}_{t-1}}{\mathcal{L}_{t-1}} \), where \(\mathcal{L}_t\) represents the training loss at epoch \({t}\). To facilitate a fair comparison of loss across consecutive epochs, we mark the data points sampled in a batch during the initial epoch and then use these same data points for numerical updates in the following epoch. Subsequently, we compare the model’s predictions $A$ with the ground truth answers $A^*$ and identify instances in which the model’s predictions are incorrect. Among these misclassified examples, we further isolate those whose loss change ratio lies within a pre-specified threshold $k$ from zero. Such samples derive minimal benefit from the training data, presumably due to the absence of novel factual information. However, the pool of these samples can still benefit from contextual redirection via symbolic updates. Consequently, we apply symbolic updates, via a call to TextGrad, specifically to a fraction $p$ of these samples to update the question $Q$ and indirectly refine internal representations and guide the model out of the local minima. We can also apply updates to system prompt $P$ in addition to updating Q. We outline the overall sequence of steps for our framework in Algorithm \ref{alg: neurosymbolic_lora}.

Another signal to monitor the saturation of numerical updates is the magnitude of the gradient norm of the training loss $ C(x)_2 = || \nabla \mathcal{L}_t ||$. The intuition behind its selection as a monitoring signal is centered at the correlation between the magnitude of the gradient norm and the effective learning of new factual knowledge. In particular, a high magnitude of the gradient norm implies significant changes in the model weights, indicating that the training sample or batch of samples potentially contained new factual knowledge that the model was otherwise not trained on. Thus, the criteria to identify saturation during numerical updates would be to monitor the magnitude of the gradient norm until it plummets below a pre-specified threshold, acting as a trigger to initiate symbolic updates in an attempt to perform context redirection given limited knowledge gain.

\subsection{Reward-based Classifier}
% Besides the criteria based on the zero-order and first-order of the loss function of the target model, 
The criteria discussed in Section \ref{sec: monitoring_signals} are based on the zero-order and first-order aspects of the loss function, meaning that they are not only sample-dependent but also dependent on the model and the training stage at which the criteria are evaluated. In contrast, content-centric evaluation and style alignment should primarily focus on the characteristics of the samples themselves. To better address this distinction, we develop a classifier that leverages a reward model to determine whether a particular sample is suited to be consumed under the numerical or the symbolic update phase. This approach ensures that the decision to perform symbolic updates is more closely aligned with the inherent properties of each sample, rather than being influenced by the dependencies of the model.
Using a fixed-reward model also enhances the ability to generalize across different datasets, enabling the identification of samples that are appropriate for symbolic updates.

% To train the reward model, we follow the standard procedure of employing pairwise preferences in reinforcement learning, as detailed in \cite{ouyang2022training}. The model should accommodate additional symbolic optimization requirements.
We examine the following reward model training approach.
%Notably, the smaller the gains from numerical optimization, the stronger the impetus for pursuing symbolic methods. Likewise, the larger the discrepancy between a sample and its symbolically optimized counterpart, the more justified the use of symbolic optimization becomes.
We initiate the process by running neurosymbolic LoRA while applying the previously mentioned criteria, such as the loss change ratio. These criteria are related to improving numerical optimization, where lower values mean less gain.
After evaluating these criteria for all samples, we construct preference pairs ${(Q^+, A^+), (Q^-, A^-)}$ by designating the sample with the higher criteria value as the positive instance $(Q^+, A^+)$ and the sample with the lower criteria value as the negative instance $(Q^-, A^-)$. Higher rewards indicate that the samples are similar to those that are symbolically optimized.

With these preference pairs established, we then proceed to minimize the loss:
\vspace{-1em}
% \textcolor{orange}{version 2 sampling }
% After evaluating 

$$\min_\theta \quad \mathbb{E}\quad \sigma\left(r_\theta(Q^+, A^+)-r_\theta(Q^-, A^-)\right),$$
where $\sigma(\cdot)$ means the Sigmoid function. It is important to note that, unlike traditional reward model training, the samples within a preference pair in our approach may originate from different questions because our objective is to evaluate and label the entire sample rather than focusing solely on individual answers.

\section{Experimental Results}
\label{sec: results}
\subsection{Experimental Setup}
\label{sec: experimental_setup}
% Math, GSm8K, AIMO, Lamp, AFLWorld 

%For generalization reasoning tasks, we evaluate on the ARC-AGi dataset \cite{chollet2019measure}, a collection of extremely challenging few-shot reasoning problems, and for personalization tasks, we evaluate on the LAMP dataset \cite{salemi2023lamp}.
\paragraph{Datasets}
We conduct several experiments across tasks such as mathematical reasoning, scientific factual claims, and textual understanding and reasoning. We utilize the training and testing sets from the GSM8K, CliniFact, and bAbi \cite{GSMdata, zhang2025dataset, weston2015towards} datasets.

\paragraph{Baselines and models} We compare the performance of our framework with two alternative baseline frameworks: purely LoRA and prefix-tuning \cite{hu2021lora,li2021prefix}. We test our framework on both SLMs and LLMs such as \text{{Deepseek-R1 Distill-Qwen-1.5B}}~\cite{guo2025deepseek} and \text{{Llama-3.1-8B-Instruct}}~\cite{dubey2024llama}.

\paragraph{Numerical update} We perform numerical updates using the Llama-3.1-8B-Instruct~\cite{dubey2024llama} and the \text{{Deepseek-R1-Distill-Qwen-1.5B}}~\cite{guo2025deepseek} models with 128 tokens for GSM8K and CliniFact and 2048 tokens for the bAbi dataset, and  set the learning rate to $5\times 10^{-6}$ and the batch size to $8$. We train for a maximum of $10$ epochs on 8 NVIDIA A4500 GPUs for these tasks which take an average of 5 hours of training time.

% Threshold |loss| >= 1 (highest loss) and <=0.1 (lowest loss) 
 
\paragraph{Symbolic update} We perform symbolic updates using calls to TextGrad with GPT-4o as the teacher model and perform at most 3 iterations.

\begin{table*}[!t]
\centering
\scriptsize
\resizebox{\textwidth}{!}{%  % \textwidth is correct here
\begin{tabular}{l*{5}{c}}
\hline
\textbf{Dataset}   & \multicolumn{2}{c}{\textbf{GSM8K}} 
                   & \multicolumn{2}{c}{\textbf{CliniFact}} 
                   & \textbf{bAbi} \\ \hline
\textbf{Model}     & \textbf{Llama-3.1-8B-Ins} & \textbf{Deepseek-R1-1.5B} 
                   & \textbf{Llama-3.1-8B-Ins} & \textbf{Deepseek-R1-1.5B} 
                   & \textbf{Deepseek-R1-1.5B}  \\ \hline
Pre-trained       & 71.56 & 9.2
                           & 58.32 & 35.54
                           & 13.7 \\
LoRA              & 74.45 & 16.9
                           & 63.21 & 37.18
                           & 14.3 \\
\textbf{NS LoRA (3)}       & \textbf{80.43} & \textbf{42.2} 
                           & \textbf{67.86} & \textbf{39.33}
                           & \textbf{15.0} \\ \hline
\end{tabular}%
}
\caption{Performance evaluation (\% accuracy) of baselines and different configurations of our framework on three datasets: GSM8K, CliniFact, and bAbi, for the \textbf{Llama-3.1-8B-Instruct} and \textbf{Deepseek-R1-1.5B} models.}
\label{tab:performance_comparison}
% \vspace{-25pt}
\end{table*}
\paragraph{Switching criteria} We utilize several distinct selection criteria to switch between numerical and symbolic updates for our analysis. We first (1) 
 randomly select a subset of the training data amounting to 10\% of the batch size and perform symbolic updates, i.e. update each question $Q$ in the subset using a call to TextGrad, at every epoch. Secondly, (2) we use the gradient norm of the training loss at every epoch as the selection criteria and update each $Q$ from a subset of the training data amounting to 10\% of the batch size with the least gradient norm. Thirdly, (3) we use the loss change ratio at every epoch as the selection criteria and update each $Q$ from a subset of the training data amounting to 10\% of the batch size with the least loss change ratio. From these selected samples, we identify instances where the model generates incorrect responses and apply symbolic updates to the corresponding prompts to enhance performance.

\subsection{Results and Discussion}
\label{sec: discussion}
To demonstrate the effectiveness of our approach, we conduct our experiments on GSM8K, Clinifact, and bAbi \cite{GSMdata, zhang2025dataset, weston2015towards} datasets. We train our model on the original training sets for 10 epochs, using a learning rate of $5\times10^{-6}$. We apply symbolic optimization to the lowest $10\%$ of the training data based on specific criteria described in Section \ref{sec: experimental_setup} for each epoch.

The evaluation is performed via answer extraction post-generation and subsequent verification of an exact match with the ground truth. This setup is inspired by a state-of-the-art evaluation harness \cite{eval-harness} with zero-shot prompting and modified instruction templates specific to both models. However, we also note that applying correct chat templates for each model can significantly affect the evaluation results, which \cite{eval-harness} does not perform for certain models. We outline the results for the baselines as well as our framework in Table \ref{tab:performance_comparison}. 

We find that implementing symbolic optimization to only $10\%$ of the batch for each epoch is enough to attain $\textbf{6.0\%}$ improvement in accuracy compared to LoRA for GSM8k for the Llama-3.1-8B-Ins model and \textbf{4.6\%} for CliniFact demonstrating the potential of this targeted strategy. For the Deepseek-R1-1.5 model this enhancement was \textbf{25.3\%} for GSM8K, \textbf{2.1\%} for CliniFact and \textbf{0.7\%} for bAbi.

\paragraph{Comparison with baselines} Per the results on GSM8K in Figure \ref{fig: acc_vs_epochs} and Table \ref{tab:performance_comparison_v2}, solely performing numerical updates using LoRA achieves poor results (74.45\% accuracy) for \text{{Llama-3.1-8B-Instruct}}. In contrast, prefix-tuning offers nearly a \textbf{3\%} increase in accuracy while the reward classifier offers a \textbf{6\%} increase in accuracy. However, our results demonstrate that pure numerical or pure symbolic updates yield inferior results compared to our neurosymbolic framework. In particular, neurosymbolic (NS) LoRA performs better than both LoRA and prefix-tuning baselines with the loss change ratio switching criteria (NS LoRA (3)) yields over \textbf{80\%} accuracy which is a \textbf{6\%} enhancement in accuracy. On the other hand, the reward-based classifier achieves comparable results with NS LoRA and is a viable alternative method. Nonetheless, we demonstrate that our hybrid approach yields performance gains for both models.

% \end{wrapfigure}

\subsection*{Experimental Details - GSM8K}
We conducted experiments in which the \text{DeepSeek-R1-} \text{Distill-Qwen-1.5B} and \text{Phi-3-mini-4k-Ins} small language models (SLMs) were fine-tuned on the GSM8K dataset using both standard LoRA and neurosymbolic LoRA approaches. All models were trained for six epochs. For neurosymbolic LoRA, the symbolic update was invoked every two epochs, with a symbolic update ratio of 10\%.

\begin{table}[h!]
\centering
\resizebox{\columnwidth}{!}{%
\begin{tabular}{lcc}
\hline
\textbf{Dataset}   & \multicolumn{2}{c}{\textbf{GSM8K}} \\ \hline
\textbf{Model}                   & \textbf{Llama-3.1-8B-Ins} & \textbf{Phi-3-mini-4k-Ins}        \\ \hline
LoRA       & 74.45                 & 80.60                                                          \\
Prefix-tuning       &  77.79                & 79.23                                                            \\
\textbf{Reward Classifier}       & 79.76                 & \textbf{81.96}                                                            \\
\textbf{NS LoRA (1)} & 78.77                     & 79.73                                                            \\
\textbf{NS LoRA (2)} & 79.75                     & 81.45                                                           \\
\textbf{NS LoRA (3)} & \textbf{80.43}                & 80.61                                                         \\ \hline
\end{tabular}%
}
\vspace{-0.5em}
\caption{Performance evaluation of baselines and different configurations of our framework on GSM8K for the \text{{Llama-3.1-8B-Instruct}} and \text{{Phi-3-mini-4k-Instruct}} models. Key methods include LoRA, prefix-finetuning, and Neurosymbolic LoRA (NS LoRA) with 3 different selection criteria: (1) a randomly selected subset forming 10\% of training data, (2) the 10\% of samples with the lowest gradient norm, and (3) the 10\% of samples with the lowest loss rate change.}
\label{tab:performance_comparison_v2}
\end{table}

% \begin{wrapfigure}{R}{0.5\textwidth}
\begin{minipage}{0.40\textwidth}
\begin{figure}[H]
    \vspace{-2.5em}
    \centering
    \includegraphics[width=\linewidth]{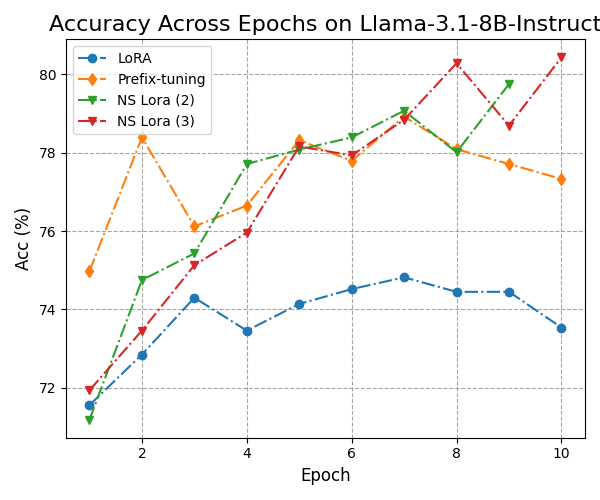}
    \vspace{-2em}
    \caption{A depiction of the trend for accuracy on GSM8k for the Llama-3.1-8B-Instruct model as the number of training epochs for the baselines and our hybrid framework NS LoRA with two primary selection criteria $C(x)_2$ and $C(x)_1$ corresponding to NS LoRA (2) and (3) respectively.}
    \vspace{0.5em}
    \label{fig: acc_vs_epochs}
\end{figure}
\end{minipage}
% \end{wrapfigure}

\begin{figure*}[h]
    \centering  
    \includegraphics[width=0.49\linewidth]{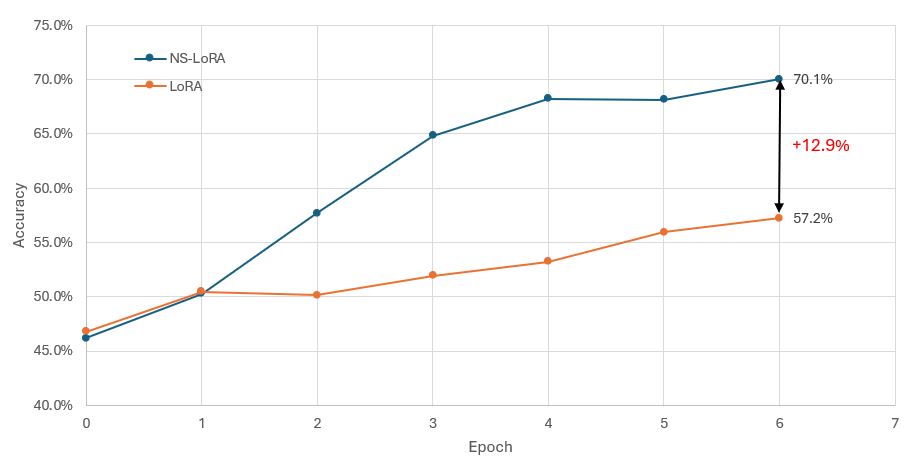}
    \includegraphics[width=0.49\linewidth]{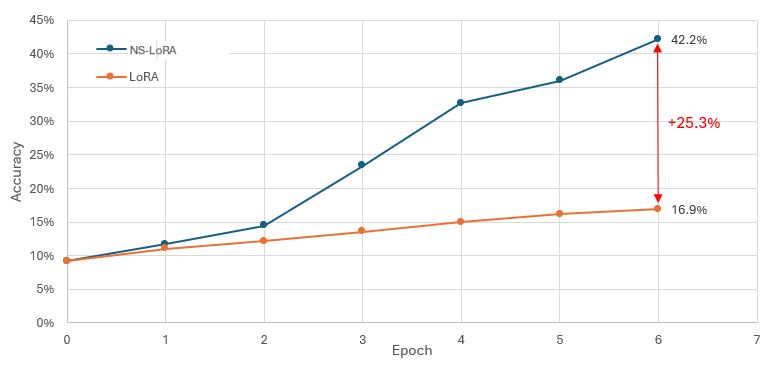}
    \caption{Performance comparison of traditional LoRA and neurosymbolic LoRA during training of the \text{Phi-3-mini-4k-Ins} and \text{DeepSeek-R1-Distill-Qwen-1.5B} model on GSM8K.}
    % \label{fig:gsm8k-phi3}
    % \vspace{-5pt}
    % \caption{Performance comparison of traditional LoRA and neurosymbolic LoRA during training on the \text{DeepSeek-R1-Distill-Qwen-1.5B} model on GSM8K.}
    \label{fig:gsm8k}
    % \vspace{-5pt}
\end{figure*}

We perform evaluation using the \text{lm-evaluation-harness} framework with the \text{gsm8k\_cot\_zeroshot} configuration and the \text{flexible-extract} filter. We set the output token length limit to the default value of 256.
Figure~\ref{fig:gsm8k} depicts the accuracy of traditional LoRA and neurosymbolic LoRA after each epoch of training of \text{Phi-3-mini-4k-Ins} and \text{DeepSeek-R1-Distill-Qwen-1.5B}.

The performance of both models show similar trends. Notably, the accuracies of both approaches remain nearly identical during the first two epochs, as symbolic updates are not performed during this phase. However, starting from the third epoch, when the symbolic update is first performed, we see an increase in the performance of our approach. While the accuracy of traditional LoRA either plateaus or improves marginally, the accuracy of neurosymbolic LoRA exhibits a significant and consistent increase.

Our results confirm that knowledge remains displaced rather than deleted and that performance can be enhanced using our hybrid framework. In addition, the results confirm that the reward-based classifier successfully captures a context-sensitive decision boundary by identifying whether the input sample requires content-level updates or style alignment. 

Given that the reported gains were realized with an upper limit of 10 epochs, our hybrid approach remains applicable for edge device deployment and in memory-constrained environments. It offers the efficiency benefits of LoRA while making sparse calls to externally-hosted TextGrad.

\paragraph{Effect of updating system prompt along with the specific question prompt }
We observe that jointly updating the system prompt and the question prompt yields significant gains across benchmarks. In particular, when paired with neurosymbolic LoRA, this strategy delivers a 3.5 \% accuracy boost on GSM8K and a 4.5 \% gain on CliniFact relative to the LoRA + system-prompt update baseline. In this setting, we first fine-tune the model via LoRA, then perform three epochs of system-prompt refinement over the full training corpus. As detailed in Table \ref{tab:system_prompt}, NS LoRA augmented with system-prompt updates consistently outperforms the simple LoRA + system-prompt combination.

Our results confirm that knowledge remains displaced rather than deleted and that performance can be enhanced using our hybrid framework. In addition, the results confirm that the reward-based classifier successfully captures a context-sensitive decision boundary by identifying whether the input sample requires content-level updates or style alignment.

% \begin{wrapfigure}{R}{0.50\textwidth}
% \begin{minipage}{0.45\textwidth}
\vspace{-0.5em}
\begin{table}[H]
\resizebox{\columnwidth}{!}{%
\centering
\scriptsize % switch to scriptsize if need be
\begin{tabular}{l*{5}{c}}
\hline
\textbf{Dataset}   & \textbf{GSM8K}
                   & \textbf{CliniFact}  \\ \hline
\textbf{NS LoRA  (3)} & 80.43 & 67.86 \\
\textbf{LoRA+Sys. Prmt. Update} & 79.71 & 65.87 \\
\textbf{NS LoRA+Sys. Prmt. Update}       & 83.24 & 70.31 \\ \hline
\end{tabular}%
% \vspace{1em}
}
\caption{Performance evaluation (\% accuracy) of models in different configurations on GSM8K and CliniFact for \text{Llama-3.1-8B-Instruct}.}
\vspace{-1em}
\label{tab:system_prompt}
\end{table}%
% \end{minipage}

Given that the reported gains were realized with an upper limit of 10 epochs our hybrid approach remains applicable for edge device deployment and in memory-constrained environments. It offers the efficiency benefits of LoRA while making sparse calls to externally hosted TextGrad.

% \subsection{Results for Small Language Models Using LM Evaluation Harness}
% \label{sec: slm_results}
% We investigate the performance of our approach using the top two open-source SLMs: \text{{DeepSeek-R1-Distill-Qwen-1.5B}} \cite{guo2025deepseek} and \text{{Phi-3-mini-4k-Instruct}} per public language model leaderboards to further evaluate our technique. 
% We limit our training epochs to 6 for each model with other parameters identical to Section \ref{sec: experimental_setup}. We compare the performance of the pre-trained models with zero-shot prompting along with LoRA and NS LoRA with the selection criteria set to $C(x)_2$. We summarize these results in Table\ref{tab:performance_comparison_slm}
% Compared to LoRA, neurosymbolic LoRA shows an accuracy improvement of about \textbf{13\%} and \textbf{26\%} for the two models, respectively. While we note that the accuracies reported are relatively low overall compared to that reported by the respective works, this is attributed to the default chat template used in LM Evaluation Harness during evaluation \cite{eval-harness}. 
% Regardless of the difference in nominal accuracy values, it is important to note the relative gain in accuracies for SLMs.
\begin{figure*}[!t]
    \centering
\includegraphics[width=\textwidth]{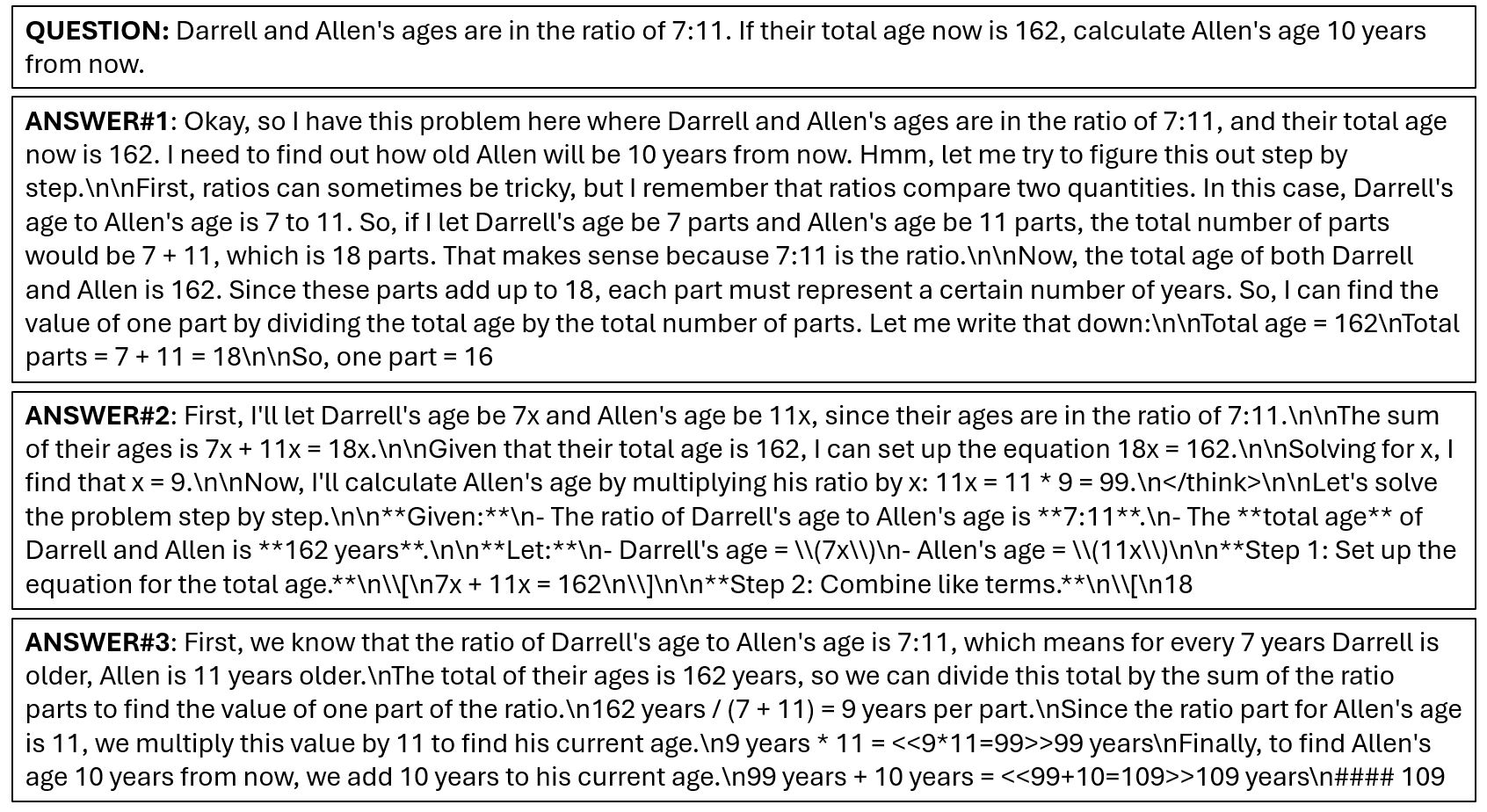}  % \textwidth instead of \linewidth
    \caption{Impact of neurosymbolic LoRA on LLM response. Answer\#1 is the response of the pre-trained LLM. Answer\#2 is the response of the LLM post LoRA. Answer\#3 is the response of the LLM post NS LoRA (3).}
    \label{fig:example}
\end{figure*}

\subsection{Higher-quality Training Set Synthesis for Downstream Fine-tuning}
Prompt editing via symbolic updates provides a means for the synthesis of high-quality training data. With symbolic data augmentation alone, downstream model performance post-fine-tuning can be substantially enhanced. To demonstrate this capability, we applied neurosymbolic LoRA on GSM8K with loss change ratio as the selection criteria on the Llama-3.1-8B-Instruct model and obtained a new augmented dataset with questions revised during the symbolic update phase. Subsequently, we incorporate the synthesized dataset to fine-tune the Qwen2.5-7B-Instruct model using LoRA. Both models were trained with a maximum training epoch limit of 5 epochs, yet converged after 2--3 epochs on average.

The experimental results show that the model attains an accuracy of 70.36\% on the original GSM8k dataset. On the augmented data, however, the model attains an accuracy of \textbf{75.96\%}. This demonstrates not only that our neurosymbolic framework for data augmentation can improve performance by over $\textbf{5\%}$ on GSM8k under zero-shot prompting, but also that the generated dataset can be consumed for downstream fine-tuning and beyond by other models. These results underscore the transferability and reusability of the synthesized dataset obtained via symbolic updates to other models. The results also demonstrate the potential of neurosymbolic scaling in data-scarce domains.
% Please add the following required packages to your document preamble:
% \usepackage{booktabs}
% \begin{table}[h]
% \resizebox{0.45\textwidth}{!}{%
% \begin{tabular}{@{}llll@{}}
% \toprule
% Dataset                                       & GSM8K & ARC-AGI & LAMP \\ \midrule
% \multicolumn{1}{l|}{Original dataset}         & 70.36 &         &      \\
% \multicolumn{1}{l|}{Augmented dataset} & \textbf{75.96} &         &      \\ \bottomrule
% \end{tabular}
% }
% \caption{The augmented dataset using symbolic updates is obtained via executing neurosymbolic LoRA with the LLAMA 3.1-8B-Instruct model and subsequently we fine-tune the QWen2.5-7B-Instruct model on the resulting dataset.The results presented correspond to the epoch with the highest accuracy.}
% \label{tab:data_transfibility}
% \end{table}
\paragraph{Impact of neurosymbolic LoRA on LLM response}
Neurosymbolic LoRA training appears to influence the response style and format of the LLM. We illustrate this effect in Figure~\ref{fig:example}, which compares outputs from three model variants: the pre-trained LLM, the LLM after traditional LoRA training, and the LLM after neurosymbolic LoRA training. All outputs are generated with a maximum token limit of 256, consistent with the default configuration in the lm-evaluation-harness framework.
The first response, from the pre-trained model, is verbose and predominantly textual, reflecting a slow and meandering reasoning process. The second response, generated after traditional LoRA training, is more succinct and incorporates more mathematical symbols and calculations. However, the reasoning halts prematurely, failing to reach a complete conclusion. In contrast, the third response, produced after neurosymbolic LoRA training, is concise, well-structured, and exhibits a coherent reasoning chain that efficiently leads to the correct answer.
These observations suggest that neurosymbolic LoRA is more effective than traditional LoRA in enabling the LLM to internalize and produce well-structured response formats and reasoning styles.

\section{Conclusion}

We demonstrate the efficacy of integrating numerical and symbolic updates within a unified framework, neurosymbolic LoRA, to enhance the performance of language models. We design a hybrid framework that performs numerical updates for factual content-level learning and symbolic updates for precise style-level adjustments. Our method showcases the strengths of synergistically leveraging both update modes. Beyond performance improvements, our approach remains memory-efficient and deployable on edge devices by sparsely making external API calls for symbolic updates and thereby optimizing computational resources. This capability is especially beneficial in data-scarce domains, such as mathematical reasoning, as it enables the generation of refined, high-quality prompts as reusable training data. Our experiments consistently demonstrate the effectiveness of a unified, reward-based signal for orchestrating numerical and symbolic update phases, and confirm that our framework outperforms purely numerical, purely symbolic, and naively combined baselines. Consequently, our findings will enable the efficient and scalable development of hybrid frameworks for fine-tuning language models. These results underscore promising future directions centered at exploring other unified monitoring signals to optimize the interplay between numerical and symbolic learning.

\paragraph{Limitations and future work} Neurosymbolic LoRA yields significant performance improvements on GSM8K across several LLMs, but shows smaller gains on other datasets. This discrepancy may be attributed to the detailed, step-by-step answers in GSM8K, which likely aids in generating targeted and instructional prompts. NS LoRA does not yet guarantee semantic preservation of rewritten prompts, and additional mechanisms for filtering or verifying symbolic edits would be needed to ensure robustness.

\bibliography{ref}

\clearpage
\onecolumn
\appendix

\section*{A \, Comparing Neurosymbolic LoRA and Lessons Learned}

\subsection{Comparing Neurosymbolic LoRA with a Two-stage Training and Prompting Approach}
We conducted additional experiments to compare neurosymbolic LoRA with a two-stage training approach that applies prompt optimization for data augmentation prior to performing numerical updates. In this approach, an optimal prompt is first generated for the entire training dataset. This prompt is then prepended to each question in the dataset, effectively augmenting the input. The resulting augmented dataset is subsequently used for traditional LoRA fine-tuning. Our experimental results depict that neurosymbolic LoRA achieves better performance than the two-stage training strategy.

\begin{figure}[h]
    \centering
    \includegraphics[width=\linewidth]{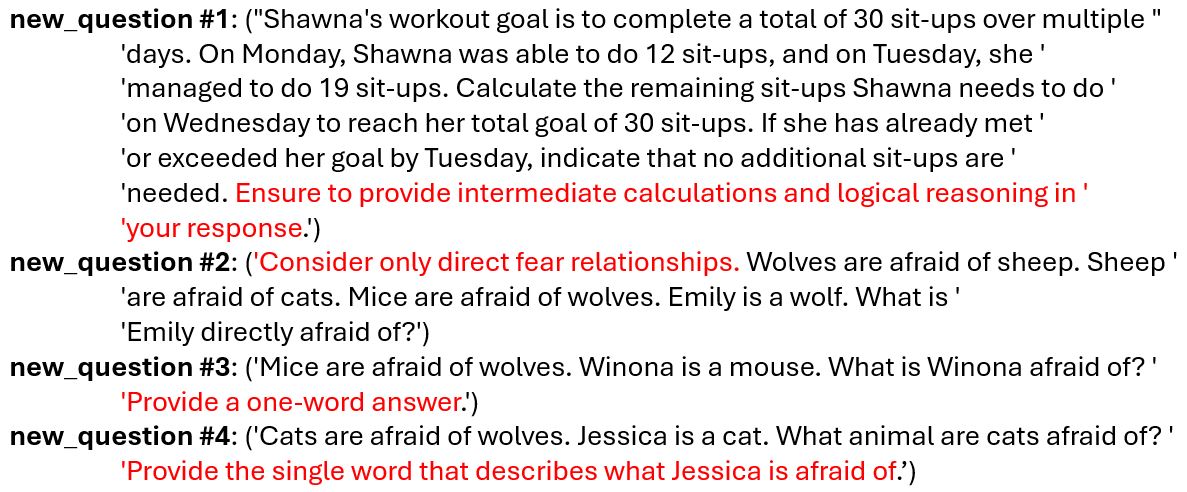}
    \caption{Examples of new questions generated by TextGrad during symbolic update. The text in red were added to the original question.}
    \label{fig:newQuestion}
\end{figure}

\subsection{Learnings from Training}
During development and experimentation, we observed that our approach yielded insignificant results on certain datasets, including BoolQ, bAbI, and OpenBookQA. A key factor contributing to this inconsistency appears to be the variability in the quality of new questions generated by TextGrad during the symbolic update step. As illustrated in Figure~\ref{fig:newQuestion}, the nature of the generated content differs significantly across datasets. In the first two examples, the added text is more instructional in nature, providing explicit guidance or reasoning strategies, which can help the model arrive at the correct answer. In contrast, the last two examples feature additions that focus primarily on the final answer rather than the reasoning process, making them less effective in deriving the correct answer. We hypothesize that this discrepancy may stem from differences in the level of detail provided in the ground truth answers, which serve as input to TextGrad when generating new questions.

\end{document}